\documentclass[letterpaper, 10 pt, conference]{ieeeconf}  

\IEEEoverridecommandlockouts                              

\usepackage{hyperref}       
\usepackage{url}            
\usepackage{booktabs}       
\usepackage{amsfonts}       
\usepackage{nicefrac}       
\usepackage{microtype}      
\usepackage{lipsum}
\usepackage{graphics} 
\usepackage{times} 
\usepackage{amsmath} 
\usepackage{amssymb}  
\usepackage{bm}  
\usepackage{algpseudocode,algorithm,algorithmicx}
\usepackage[shortcuts,acronym]{glossaries}
\usepackage{gensymb} 
\usepackage{caption}
\usepackage{graphicx}
\usepackage{multirow}
\usepackage{array}
\usepackage{subcaption}
\usepackage{xcolor}
\usepackage[outdir=./]{epstopdf}
\newcolumntype{C}[1]{>{\centering}m{#1}}

\title{Multi-UAV Coverage Path Planning for the Inspection of Large and Complex Structures}

\author{Wei Jing$^{1,2,4}$, Di Deng$^{3}$, Yan Wu$^{1,2,*}$ and Kenji Shimada$^{3}$
\thanks{$^{1}$ A*STAR Institute of Information Research (I$^2$R), Singapore. }%
\thanks{$^{2}$ A*STAR Initiative for Artificial Intelligence and Analytics (AI$^{3}$), Singapore. }%
\thanks{$^{3}$ Department of Mechanical Engineering, Carnegie Mellon University, Pittsburgh, PA, USA. }%
\thanks{$^{4}$ A*STAR Institute of High Performance Computing (IHPC), Singapore. }%
\thanks{$^{*}$ Corresponding Author, Email: {\tt\small wuy@i2r.a-star.edu.sg}}%
}

\newacronym{rkga}{RKGA}{Random Key Genetic Algorithm}
\newacronym{brkga}{BRKGA}{Biased Random Key Genetic Algorithm}

\newacronym{cpp}{CPP}{Coverage Path Planning}
\newacronym{sc-vrp}{SC-VRP}{Set-Covering Vehicle Routing Problem}
\newacronym{scp}{SCP}{Set Covering Problem}
\newacronym{vrp}{VRP}{Vehicle Routing Problem}

\newacronym{uav}{UAV}{Unmanned Aerial Vehicle}

\begin{document}
\maketitle


\begin{abstract}
 We present a multi-UAV \gls{cpp} framework for the inspection of large-scale, complex 3D structures. In the proposed sampling-based coverage path planning method, we formulate the multi-UAV inspection applications as a multi-agent coverage path planning problem. By combining two NP-hard problems: \gls{scp} and \gls{vrp}, a \gls{sc-vrp} is formulated and subsequently solved by a modified \gls{brkga} with novel, efficient encoding strategies and local improvement heuristics. We test our proposed method for several complex 3D structures with the 3D model extracted from OpenStreetMap. The proposed method outperforms previous methods, by reducing the length of the planned inspection path by up to 48\%.
\end{abstract}


\section{Introduction}

\noindent While visual inspection of 3D structures with a camera-equipped \gls{uav} can be achieved by a manual flight, it becomes extremely difficult for a human operator to perform the task efficiently and safely for a large-scale, complex 3D structure.  It is particularly difficult for a human operator to make sure that the camera captures every bit of the target surfaces while flying the UAV safely in a windy condition. For complex 3D structures, it is thus necessary to fly a \gls{uav} autonomously along optimal paths planned prior to the flight \cite{bircher2015structural, jing2016sampling, roberts2017submodular}.  By flying along carefully planned paths, we can make sure that an autonomous UAV can capture every portion of the target surfaces by photos and/or videos with shortest possible flight paths.

 If the 3D structures are not only complex but also large-scale, deploying multiple autonomous UAVs is desirable, considering that typical quad-rotor UAVs’ batteries last only for 20 - 40min.  With the recent price drop of quad-rotor UAVs, deploying multiple autonomous UAVs for visual inspection is a realistic solution for completing visual inspection tasks for a large, complex 3D structure \cite{almadhoun2019survey}. Planning optimal paths for a team of multiple UAVs is the subject of the work presented in this paper.

Planning paths for multiple UAVS is significantly more complex and challenging than in a single UAV case. In formulating the problem, we assume that the geometries of target 3D structures are known and represented as triangular meshes, made available through OpenStreetMap buildings \cite{OpenStreetMap}.  Assuming also that the geometries do not change between the planning time and the time that the UAVs are deployed, the problem becomes a model-based, offline Coverage Path Planning (CPP) problem \cite{almadhoun2016survey, scott2003view}.  Since multiple UAVS, or agents, are used, the problem is also considered as a Multi-Agent(UAV) Coverage Path Planning (MACPP) problem.

\begin{figure}[t]
    \centering
    \includegraphics[width=0.99\linewidth]{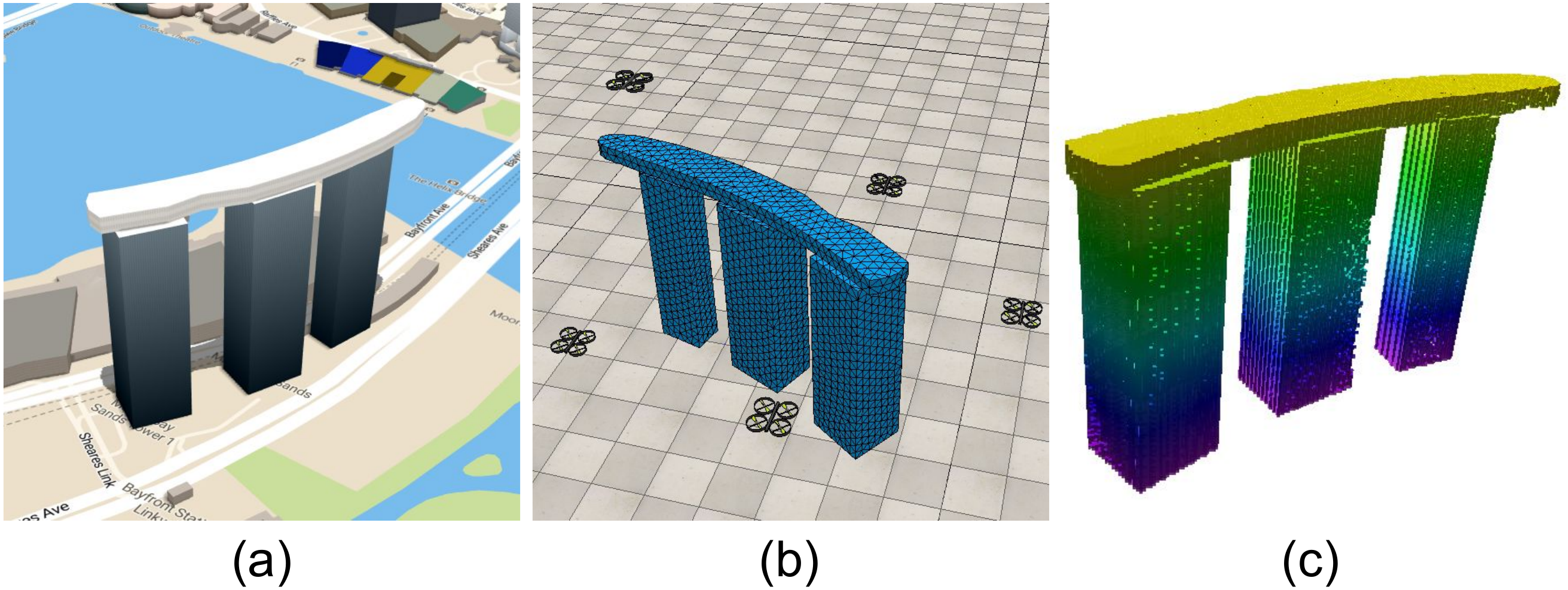}
    \caption{Large-scale complex 3D structure inspection: a): 3D model (Marina Bay Sands) from OpenStreetMap buildings \cite{OpenStreetMap}; b) multi-UAV inspection path planning; c): voxel constructed from Octomap \cite{hornung13auro}}
    \label{fig:frontpage}
    \vspace{-4mm}
\end{figure}

Previously published planning methods are not suitable for solving the multi-agent path planning for visual inspection of large and complex 3D structures. These methods are limited in that most of them are for a single agent \cite{song2017online, jing2019coverage} and that the ones for multiple agents mostly work only for 2D problems or simple 3D problems \cite{cabreira2019survey, deng2019constrained}.

This paper presents a novel formulation and algorithm for the CPP problem with multiple UAVs for 3D visual inspection tasks. The proposed approach utilizes a sampling-based method to generate and select the via-points and path-primitives. Upon evaluation of the path and the visibility, a Coverage Probabilistic Roadmap (C-PRM) graph is constructed to encode the topological, distance and coverage information. The overall problem is then formulated as a Set Covering Vehicle Routing Problem (SC-VRP) and solved by a proposed Biased Random Key Genetic Algorithm (BRKGA). The main contributions of our work are:
\begin{itemize}
    \item a multi-\gls{uav} \gls{cpp} framework for large-scale complex 3D visual inspection tasks using a sampling-based planning method;
    \item a minmax \gls{sc-vrp} formulation by combining two NP-hard problems, \gls{scp} and \gls{vrp} for the MACPP problem;
    \item an efficient evolutionary computing algorithm with novel encoding and decoding strategy, as well as local improvement heuristics to solve the formulated minmax \gls{sc-vrp} problem.
\end{itemize}

\section{Relevant Work} \label{sec:headings}


\noindent Path Planning finds a collision-free path for a robot to move from a starting pose to a target pose \cite{lavalle2006planning}. \gls{cpp} \cite{choset2001coverage, cabreira2019survey} is to plan an efficient path that covers a specified area with or without the starting and target pose requirements. \gls{cpp} problems \cite{cabreira2019survey} have been studied extensively in many applications such as inspection \cite{jing2019coverage, fu2019toward}, surveying \cite{vasquez2019coverage} and 3D mapping / reconstruction \cite{hepp2018plan3d}; for different types of robotic systems such as UAV, UGV and robotic manipulators. 

Depending on availability of the target geometric information, \gls{cpp} could be either model-based or non model-based \cite{almadhoun2019survey, scott2003view}. For \gls{cpp} in inspection applications, usually the 3D model of the target is available and could be formulated as a model-based \gls{cpp} problem \cite{almadhoun2016survey, englot2013three}; with only a few exceptions \cite{song2017online}. Sampling-based ``Generate-Test'' framework are usually used for the model-based \gls{cpp} problems \cite{scott2009model, jing2016view, englot2013three}, which usually generate redundant samples to fulfill the coverage requirements of the inspection and then optimize the path and viewpoints from the samples. Learning-based methods has also been combined with the sampling-based method for the single-agent \gls{cpp} problem \cite{kaba2017reinforcement}.

\gls{cpp} with multiple UAVs has also been studied \cite{almadhoun2019survey, galceran2013survey, deng2019constrained} to reduce the required time or to cover larger areas. Most of the work mainly focused on 2D swapping coverage with UAVs for covering 2D areas \cite{deng2019constrained, nedjati2016complete, kim2020voronoi}, with different optimization objectives depending on the requirements from the applications. Only a few studies focused on 3D coverage planning with multiple agents. However, the 3D environments in these work are limited to simple smooth targets (e.g. 3D terrain) \cite{choi2019energy, mansouri2018cooperative}. More recently, research work on 3D structural coverage with multiple agents have been conducted, but not targeting for a visual inspection application that requires a high coverage ratio \cite{renzaglia2019multi}. More comprehensive reviews of state-of-the-art planning methods for \gls{cpp} and multi-robot \gls{cpp} problems in different robotic applications are available in previous survey papers \cite{scott2003view, chen2011active, galceran2013survey, cabreira2019survey, almadhoun2019survey}. 

In this paper, we address the model-based 3D MACPP problem by a sampling-based ``Generate-Test'' framework \cite{scott2009model, tarbox1995planning}, with the path-primitive sampling method \cite{jing2019coverage} to ensure a high ratio of coverage of a large-scale, complex 3D environment. This work extends the path-primitive sample method and applies it to a multi-UAV inspection task. A novel BRKGA method is applied to solve the NP-hard optimization problem.
\section{Problem Description}

\noindent The goal of our work is to solve the path planning problem for 3D visual inspection with multi-UAV system for large-scale, complex structures. Since our solution is for an inspection task for static environment, we assume the geometry of the environment is known, and it is represented by a triangular mesh format. Thus, the planning problem is formulated as a model-based, offline planning problem. For the multi-agent system, $K$ UAVs with onboard cameras fly around the buildings to conduct the inspection. The objective is to minimize the length of the inspection path for the task, with the coverage requirement constraints. The planning problem for 3D structural inspection with multi-UAV is thus formulated as a MACPP problem.

\section{Method}

\begin{figure}[t]
    \centering
    \includegraphics[width=0.99\linewidth]{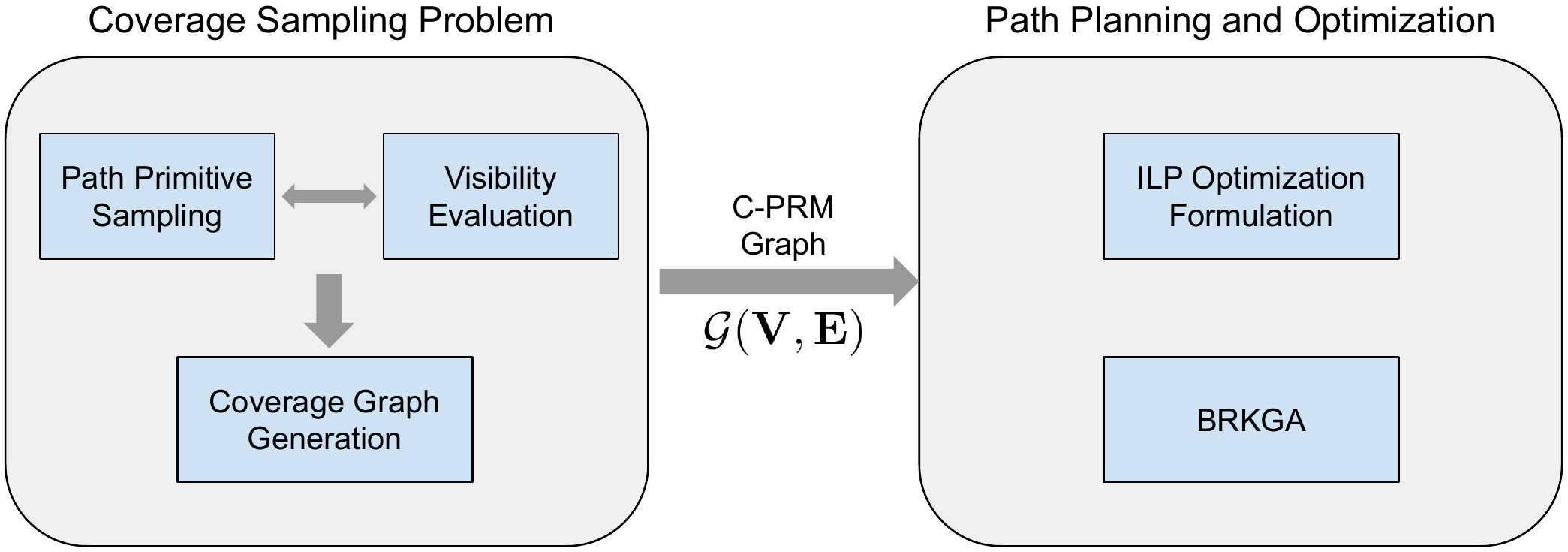}
    \captionsetup{format=hang}
    \caption{Overview of our proposed framework which breaks down the problem into a coverage sampling problem followed by a path planning and optimization problem.}
    \label{fig:overview}
    \vspace{-4mm}
\end{figure}

\noindent We propose an inspection path planning framework for an MACPP problem and apply it to the 3D visual inspection task. The proposed method first generates way-points and path-primitives by incremental sampling to create a C-PRM graph with information on topology, coverage and path length; the MACPP is then formulated as a min-max SC-VRP problem and solved by \gls{brkga}. The overall structure of the proposed problem-solving framework is shown in Fig. \ref{fig:overview}

\subsection{Coverage Sampling with Path-Primitives}

\noindent For the \gls{cpp} problem in inspection applications, a Coverage Sampling Problem (CSP) \cite{englot2013three} is formulated to generate and select the via-points and path-primitives to ensure good coverage of the target structures. 

\subsubsection{Sampling and Visibility Evaluation}

We use a path-primitive sampling method to generate and select the via-points and paths for 3D swapping coverage \cite{jing2019coverage}. In addition, we also adopt a dual sampling method \cite{englot2013three} to improve the sampling efficiency and improve the coverage. Moreover, binary dilation on voxel is also used to create efficient sampling space for the via-points and path-primitive generation in the \textit{RandomSample} and \textit{DualSample}. A binary dilation of voxel $\mathcal{A}$ by $\mathcal{B}$ is defined as:
\begin{align} \label{eqn:dilation}
    \mathcal{A} \oplus \mathcal{B} = \{x | \mathcal{\hat B}_x \cup \mathcal{A} \neq \emptyset \}
\end{align}
where $\mathcal{\hat B}_x$ shifts the symmetric of $\mathcal{B}$ by $x$.

The dual path-primitive sampling algorithm is shown in Algo \ref{algo:primitive}. It iteratively samples the viapoints and path-primitives with a bias towards unseen surface patches. In order to improve the via-point sampling efficiency, the sampling space is created by two binary dilations, $D_1$ and $D_2$, as shown in Eqn. \ref{eqn:dilation} and Algo \ref{algo:primitive}, and a subtraction. \textit{RandomSample} is then performed to sample the via-points. This is followed by iterative incremental \textit{DualSample} and visibility evaluation. The orientations of the via-points and the path primitives are generated using potential field methods and linear interpolation, details could be found in \cite{jing2019coverage}. 

The visibility of a viewpoint to the samples on the target surface is evaluated through ray-tracing with the given camera specifications \cite{scott2009model, jing2016view}. In addition to the viewpoints, \textit{VisibilityEvaluation}($\mathbf{E}, \mathbf{M}$) evaluate the visibility of path-primitives by sampling viewpoints on the path and combining the binary visibility information of the samples \cite{jing2019coverage}. The target surface is uniformly sampled into $\mathbf{M}$, a set of triangular patches for the visibility evaluation. A $m \times 1$ binary vector, $\mathbf{s}$, is computed for each path-primitive to represent the visibility of the target objects. The operator, \textit{dilate($\mathbf{M}_v, d$)}, performs a dilation on $\mathbf{M}_v$ with a sphere of radius $d$. $\mathcal{L}(v_i, v_j, \mathbf{M})$ is the collision-free local planner to find the collision-free local path from $v_i$ to $v_j$. \textit{DualSampleVP} samples the via-points with a bias towards uncovered surface patches.

\begin{algorithm}
\small
\caption{Coverage Sampling with Dual Path-Primitive Sampling}
\label{algo:primitive}
\begin{algorithmic}[1]
    \Require
        The collision-free Local Planner, $\mathcal{L}$; the 3D model of target objects in triangular mesh format $ \mathbf{M}$, and in binary voxel format $ \mathbf{M}_v$; the max viewing range of sensor, $d_{vis}$; and the safety distance, $d_{safe}$ 
    \Ensure
     		The set of sampled via-points $\mathbf{V}$; and the set of sampled path-primitives, $\mathbf{E}$. 
\State $ \mathbf{E}, \mathbf{V}, \mathbf{M}_{unseen} \leftarrow \emptyset, \emptyset, \mathbf{M}$
\State $D_1, D_2  \leftarrow $ dilate($\mathbf{M}_v, d_{vis}$), dilate($\mathbf{M}_v, d_{safe}$)
\State $D \leftarrow $ VoxelSubstract($D_1, D_2$)
\State $\mathbf{V} \leftarrow$ RandomSampleVP($D$)
\While{ $|\mathbf{M}_{unseen}| > m_{min}$ or $|\mathbf{E}| < n_{desired}$}
	\State $\mathbf{V'} \leftarrow$ DualSampleVP($D, \mathbf{M}_{unseen}$)
	\State $\mathbf{V} \leftarrow $  Append($ \mathbf{V}, \mathbf{V'}$)
    \For{$v_i, v_j \in \mathbf{V}, i \neq j$}
    \If{ dist($v_i, v_j$) $ \leq d_{max}$ }
    \State $e_{ij} \leftarrow \mathcal{L}(v_i, v_j, \mathbf{M})$ 
    \State $\mathbf{E} \leftarrow $  Append($ \mathbf{E}, e_{ij}$)
    \EndIf
    \EndFor
    \State $\mathbf{M}_{unseen} \leftarrow$ VisibilityEvaluation($\mathbf{E}, \mathbf{M}$)
\EndWhile
\State \Return{$\mathbf{V}, \mathbf{E}$}
\end{algorithmic}
\end{algorithm}

\subsubsection{Coverage Probabilistic Roadmap for Inspection Planning}

\begin{figure}[!ht]
    \centering
    \begin{subfigure}[b]{0.49\linewidth}
    	\includegraphics[width=0.99\linewidth]{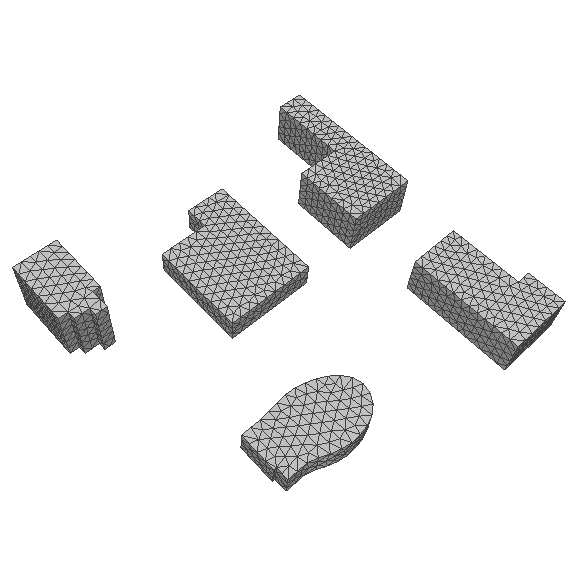}
    	\caption{Target structure}\label{fig:cprm_a}
    \end{subfigure}
    \begin{subfigure}[b]{0.49\linewidth}
    	\includegraphics[width=0.99\linewidth]{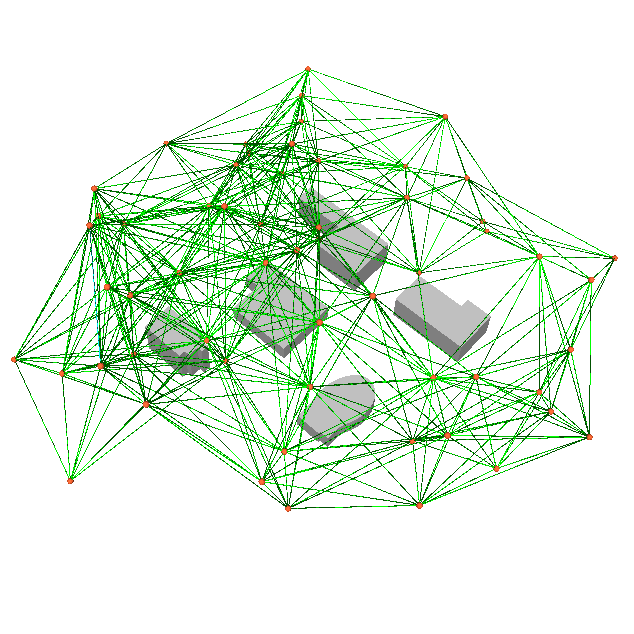}
    	\caption{Target structure with C-PRM}\label{fig:cprm_b}
    \end{subfigure}
    \caption{Illustration of C-PRM Graph}\label{fig:cprm}
    \vspace{-4mm}
\end{figure}

After coverage sampling and visibility evaluation, a Coverage Probabilistic Roadmap (C-PRM) graph, $\mathcal{G}(\mathbf{V}, \mathbf{E})$, with coverage information for inspection is formulated for a multi-UAV system. Similar to the construction of a Probabilistic Roadmap (PRM) \cite{kavraki1996probabilistic}, the C-PRM encodes additional coverage information of the path-primitives in edge $\mathbf{E}$ of $\mathcal{G}$. An illustration of sampled C-PRM is shown in Fig. \ref{fig:cprm}. Note that more via-points and path-primitives are sampled during the actual implementation.

In addition to topological information, edge $e_{ij} \in \mathbf{E}$ also encodes distance information and coverage information: $\lbrace d_{ij}, \mathbf{s}_{ij} \rbrace $. For the binary element $s_{ij}^k \in \mathbf{s}_{ij}, k\in [1, m]$, $s_{ij}^k = 1$ indicates that a surface patch indexed by $k$ is visible to path-primitive $e_{ij}$, and $s_{ij}^k = 0$ indicates that it is invisible. 

\subsection{Planning and Optimizing the Inspection Path}

\noindent The inspection path planning problem with multiple UAVs could be formulated as an SC-VRP by combining two NP-hard problems: SCP and VRP. The coverage problem is formulated by sampling the target surface and evaluating the visibility. The C-PRM graph, $\mathcal{G}(\mathbf{V}, \mathbf{E})$, is used for the formulation of the optimization problem for MACPP.

\subsubsection{Integer Linear Programming}

The Integer Linear Programming (ILP) formulation of min-max SC-VRP is given in Eqn. (\ref{eqn:minmax_obj}), with the objective of minimizing the maximum length of the individual UAV path. 

\begin{equation} \label{eqn:minmax_obj}
    \min_{\mathbf x} \max_k \sum_{i=1}^n\sum_{j=1}^m d_{i,j}\hat{x}_{i,j,k},
\end{equation}
subject to the following constraints:
\begin{align}
    &\sum^n_{i=1} \sum^K_{k=1}  \mathbf{s}_{ij} \hat{x}_{i,j,k} \geq 1 \quad \forall j, \label{eq:cov2}\\
    &\sum_{i,j | v_i, v_j \in V'_{k}} \hat{x}_{i,j,k} \leq | L |-1, \quad | L | \in {V'_{k}} , | L | \geq 2,  \label{eq:subtour}\\
    &V'_{k} | V'_{k} \in V, for \quad \hat{x}_{i,j,k} = 1, \\
    & e_{i,j} \in \mathbf{E}, for \quad \hat{x}_{i,j,k} = 1, \\
    &\hat{x}_{i,j,k} \in \{0, 1\}, 
\end{align}
where $x_{i,j,k}$ is the binary selection of the path, $e_{i,j}$, in C-PRM $\mathcal{G}(\mathbf{V}, \mathbf{E})$ of UAV $k$ (as in Constraint (6)).  Constraint (\ref{eq:cov2}) is the coverage constraints, which states all surface areas should be covered by adding coverage information $ \mathbf{s}_{ij}$ of all selected paths from all UAVs. Constraint (\ref{eq:subtour}) is the subtour elimination constraint for all $k$ UAVs of selected subset of the nodes \cite{dantzig1954solution, laporte1992traveling}. Since the nodes are via-points to provide connectivity information of edges, and the coverage is done through flying over the path-primitive (edge), there is no constraint on one-visit per node (via-point), or visiting of all nodes.

In many inspection applications, a desired coverage ratio, $\delta_d $, is specified instead of full coverage. In such a case, the coverage constraint, Constraint (\ref{eq:cov2}), should be alternated to:
 \begin{align} \label{eq:cov3} 
       \frac{ \sum^m_{j=1} (\sum^n_{i=1} \sum^n_{k=1} \mathbf{s}_{ij} \hat{x}_{i,j,k}) \geq 1 ) } {m} \geq  \delta_d 
 \end{align}
which states the coverage ratio should be larger than or equal to $\delta_d $, by summing all coverage information $ \mathbf{s}_{ij}$ of all selected paths from all UAVs.
\subsubsection{Path Planning and Optimization with BRKGA}

\gls{rkga} is a meta-heuristic framework that uses random keys in the chromosome to encode the solution \cite{bean1994genetic}. RKGA has been applied to many combinatorial optimization problems \cite{snyder2006random, gonccalves2011biased, jing2017redundant}. Using random keys is to add an encoding-decoding process to map the feasible space with complex constraints to an unconstrained one, which reduces the complexity of constraint handling in solving a constrained combinatorial optimization problem. 

For the formulated NP-hard \gls{sc-vrp}, we adopt an improved version of \gls{rkga}, \gls{brkga} \cite{gonccalves2011biased}, with a modified encoding-decoding method. The chromosome $C$ is a $n \times 1 $ vector ($n$ is the number of node in $\mathcal{G}$), which consists $n$ random-keys $\{x | 0 \leq x \leq K\}$. The integer part of the random key is used to encode the UAV (e.g. $x_{int} = 1$ means UAV $1$ is selected), and the fractional part is used to encode the neighbour of current to choose. Note that not all nodes need to be visited since this is a set covering problem, instead, the task is considered as completed as long as the required coverage is achieved. Additionally, we use \gls{brkga} with local improvement heuristics to further optimize the performance. 

Our new formulation transfers the ILP problem with many constraints into a sequencing problem. As shown in Algo. \ref{algo:rkga}, the fitness evaluation of \gls{brkga} decodes the chromosome into paths of $K$ UAVs. The coverage constraint will be evaluated after decoding each random-key. Once the coverage constraint is satisfied, the fitness evaluation exits and returns the fitness value, as well as the paths of the UAVs. 

\begin{algorithm}
\small
\caption{Fitness Evaluation}
\label{algo:rkga}
\begin{algorithmic}[1]
    \Require
      		The input chromosome, $C$; the C-PRM graph, $\mathcal{G}(\mathbf{V}, \mathbf{E})$; and other algorithm specific parameters; 
    \Ensure
     		The fitness score $s_{max}$
\State $ \mathbf{V}_{res}, \mathbf{E}_{res}, \delta, \mathbf{s} \leftarrow \emptyset, \emptyset, 0, \mathbf{0}$
\For{$i \leq$ sizeof$(C)$}
    \State $v, e, x_{int} \leftarrow$ decode($v, \mathcal{G}, c_i$)
    \State $ \mathbf{E}_{res} \leftarrow$ append($\mathbf{E}_{res},e$) 
    \State $\delta \leftarrow$ evaluateCoverage($ \mathbf{E}_{res}$) 
    \State $ s_{x_{int}} \leftarrow s_{x_{int}}+d(e)$ 
    \If{{$\delta \ge \delta_d$}}
        \State break
    \EndIf
\EndFor
\State $ s_{max} \leftarrow \mathbf{s} $ 
\State \Return{$s_{max}$}
\end{algorithmic}
\end{algorithm}

In order to further improve the performance, we also added local improvement heuristic to the \gls{brkga}. Similar to the 2-opt heuristics in Traveling Salemans Problem (TSP) \cite{croes1958method}, the local improvement heuristics shown in Algo. (\ref{algo:2-opt}) is used to find two valid vertices in the C-PRM $\mathcal{G}$ to perform the \textit{2optSwap} operation for each path of a individual UAV. \textit{2optSwap}($C_{o}, v_i, v_j$) only performs the swap operation when the path length could reduce while the coverage constraint still holds. \textit{2optSwap} also re-encodes the random-keys at the swapping points, and returns the swapped chromosome $C_o$. This process is repeated along the paths of all UAVs in $P$. For the BRKGA with local improvement heuristics, the local improvement heuristic is used as an additional operator; in addition to the conventional RKGA mutation, crossover, and selection operators.

\begin{algorithm}
\small
\caption{Local Improvement Heuristics}
\label{algo:2-opt}
\begin{algorithmic}[1]
    \Require
      		The input chromosome, $C$; the C-PRM graph, $\mathcal{G}(\mathbf{V}, \mathbf{E})$; and the inspection paths(sequence of nodes) of all UAVs, $P$; 
    \Ensure
     		  the output chromosome $ C_{o}$; 
\State $ C_{o} \leftarrow C$
\For{$p_k \in P$}
    \For{$v_i \in p_k$}
        \For{$v_j \in $ neighbour($v_i, \mathcal{G}$)}
            \State $C_{o} \leftarrow $ 2optSwap($C_{o}, v_i, v_j$)
        \EndFor
    \EndFor
\EndFor
\State \Return{$C_{o}$}
\end{algorithmic}
\end{algorithm}

\section{Implementation and Results} \label{sec:res}

\subsection{Setup}
\begin{figure}[!ht]
\centering
\begin{subfigure}[b]{0.32\linewidth}
	\includegraphics[width=0.99\linewidth]{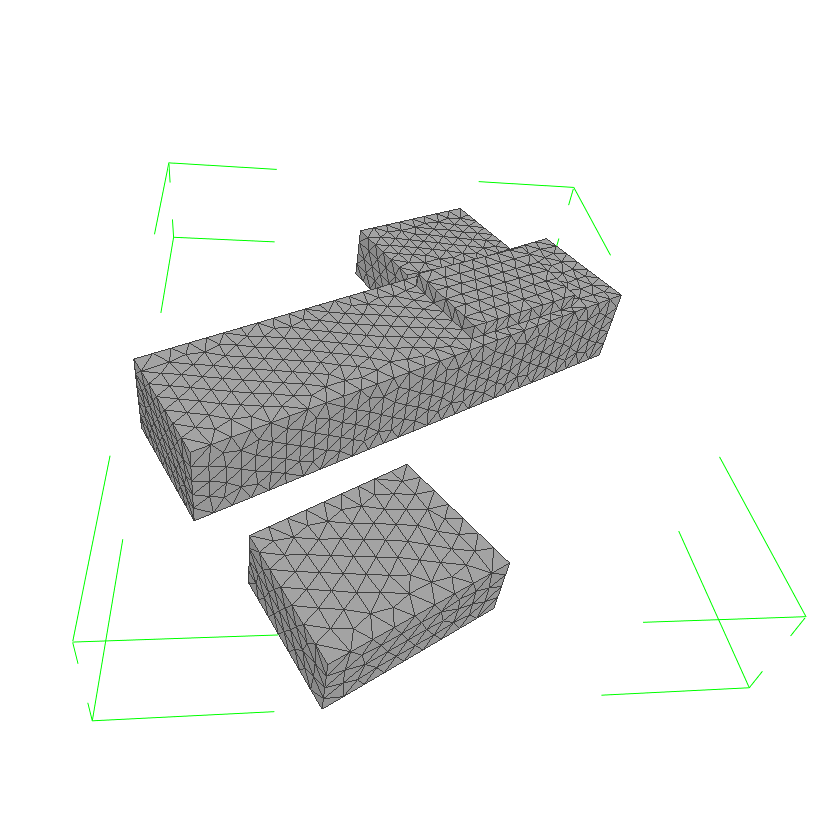}
	\caption{T1: $64\times80\times13m$}\label{fig:octo_ny7}
\end{subfigure}    
\begin{subfigure}[b]{0.32\linewidth}
	\includegraphics[width=0.99\linewidth]{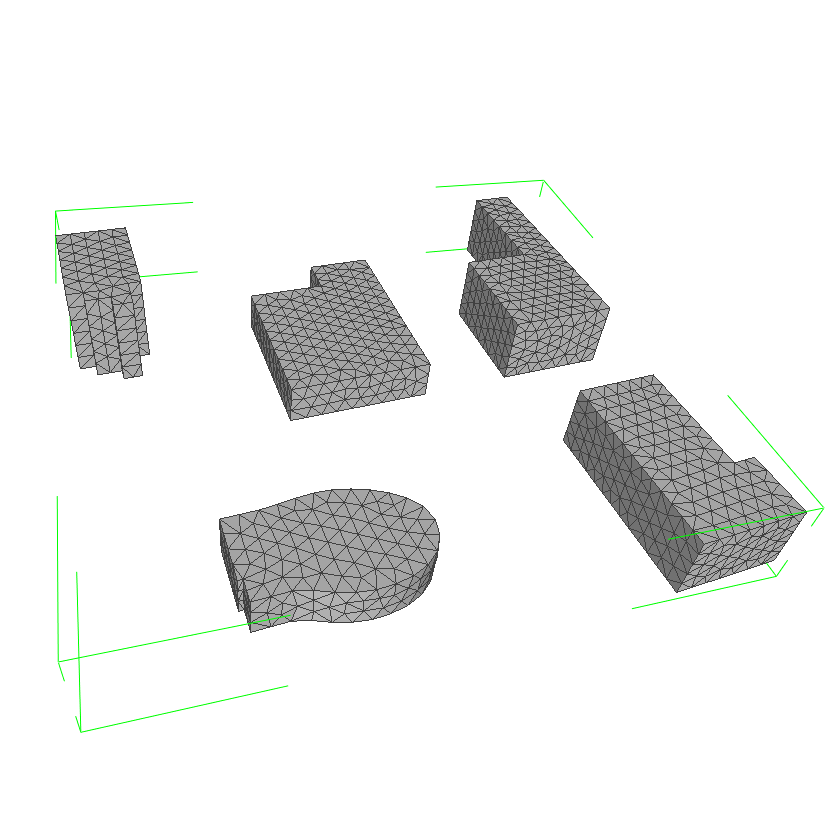}
	\caption{T2: $85\times79\times14m$}\label{fig:octo_b42}
\end{subfigure}
\begin{subfigure}[b]{0.33\linewidth}
	\includegraphics[width=0.99\linewidth]{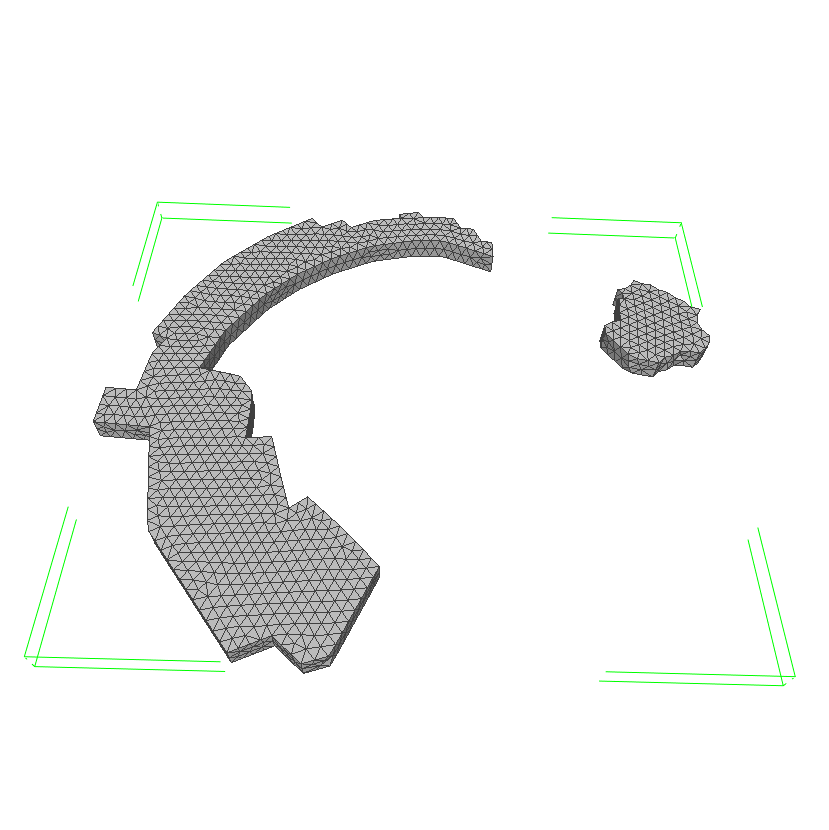}
	\caption{T3: $164\times190\times8m$}\label{fig:octo_ny7}
\end{subfigure}

\begin{subfigure}[b]{0.32\linewidth}
	\includegraphics[width=0.99\linewidth]{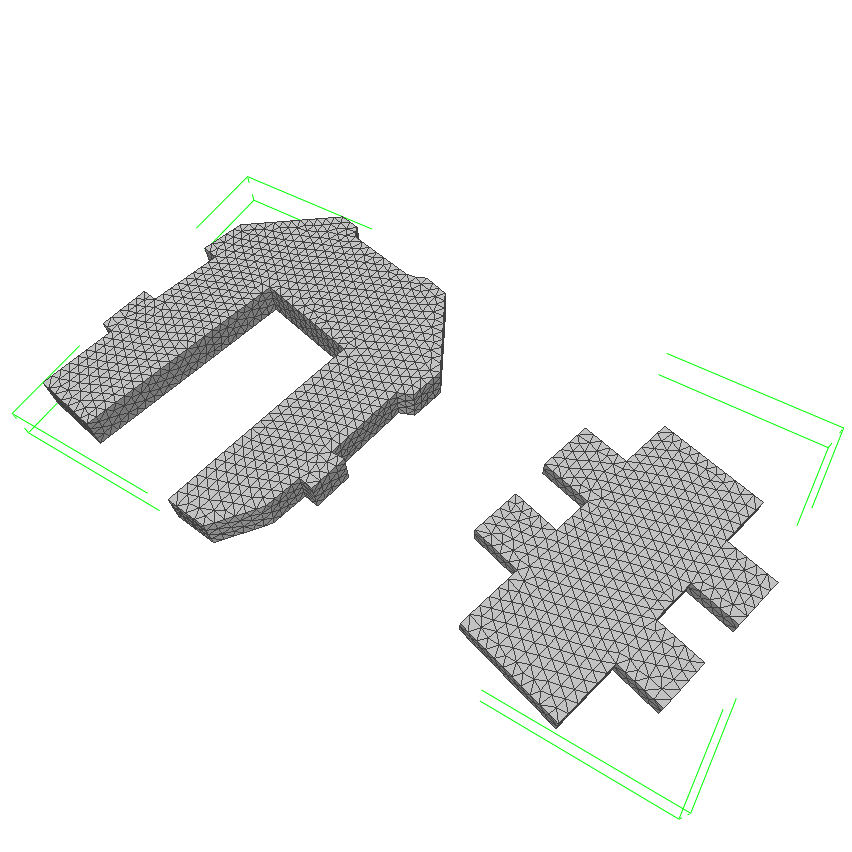}
	\caption{T4: $286\times156\times15m$}\label{fig:octo_b42}
\end{subfigure}
\begin{subfigure}[b]{0.32\linewidth}
	\includegraphics[width=0.99\linewidth]{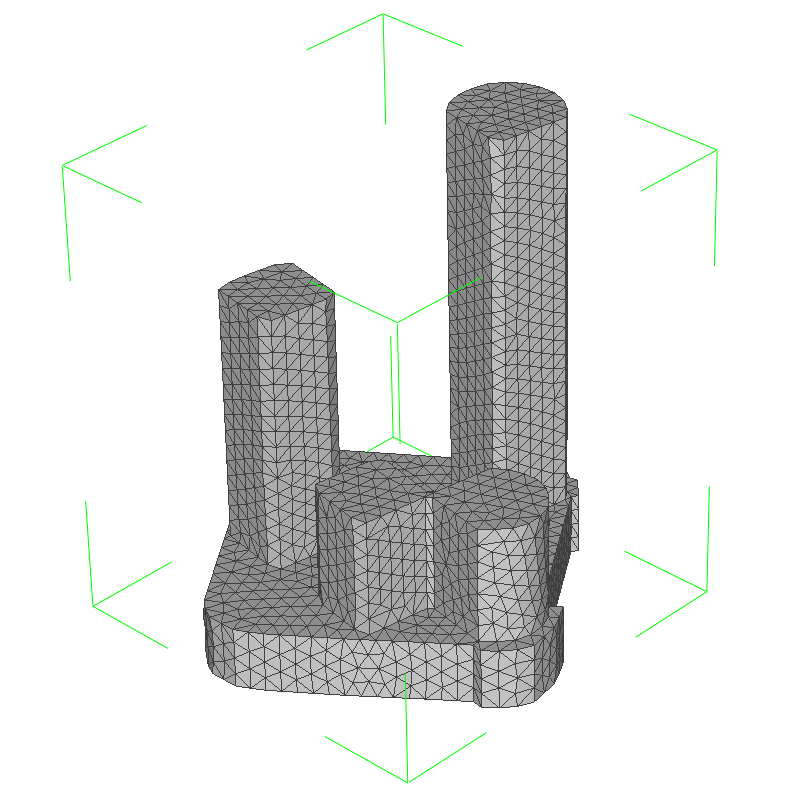}
	\caption{T5: $198\times196\times226m$}\label{fig:octo_b42}
\end{subfigure} 
\begin{subfigure}[b]{0.33\linewidth}
	\includegraphics[width=0.99\linewidth]{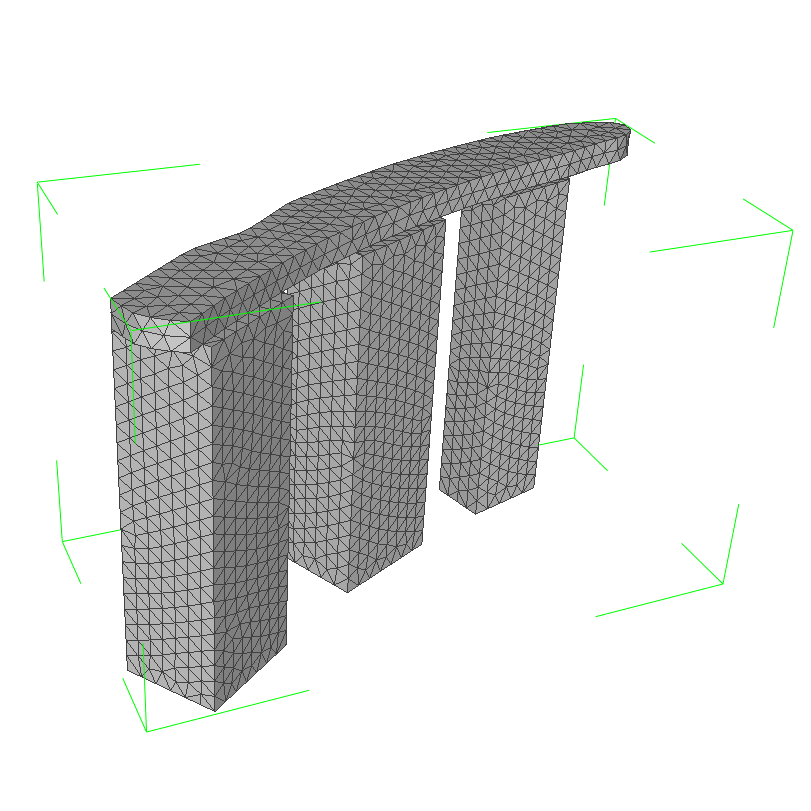}
	\caption{T6: $159\times319\times207m$}\label{fig:octo_b42}
\end{subfigure}
	\caption{Target structures and their sizes}\label{fig:models}
	\vspace{-4mm}
\end{figure}

\noindent We validate the proposed planning method with six 3D visual inspection cases. The 3D models of the target structures are real-world buildings in Singapore and Pittsburgh, taken from our previous work or extracted from the OSM Buildings of OpenStreetMap \cite{OpenStreetMap}. The models are uniformly remeshed to create samples $m \in \mathbf{M}$ on the target surface, as shown in Fig. \ref{fig:models}. The sizes of the bounding boxes are also indicated. We implemented BRKGA using DEAP \cite{DEAP_JMLR2012}, the graph $\mathcal{G}$ and relevant graph algorithms using NetworkX \cite{hagberg2008exploring}. For the parameters of visibility evaluation, the diagonal FOV is set to $94  \degree $; the maximum viewing angle is $75 \degree$; the safety distance is $2m$; the maximum viewing ranges and the coverage ratios are $50m$ and $99\%$ respectively for the first two small scale targets, and $70m$ and $98\%$ respectively for the last four large-scale targets. The parameters used are similar to those in the literature \cite{jing2016view} \cite{jing2019coverage}.

\subsection{Results of CPP for Inspection} \label{sec:res_b}

\begin{table}[!ht]
\scriptsize
\centering
\caption{Results of planned path for single/multi-agent inspection (measured in meters, average of 10 runs)} \label{table::res_1}
\begin{tabular}{ccccccc}
\toprule
\multirow{2}{*}{} & \multicolumn{3}{c}{\textbf{T1}} & \multicolumn{3}{c}{\textbf{T2}}  \\
   No. of UAVs               & 1 & 2 & 3               & 1 & 2 & 3               \\ 
\midrule
\textbf{VPP-TSP\cite{jing2016sampling, chen2004automatic}}  & $507.7$    & -  & -                 & $587.5$    & -  &    -       \\
\textbf{GNS\cite{jing2019coverage}}      & $425.6$    & $325.9$   & $313.1$    & $466.2$    &  $441.2$ &  $344.1$               \\
\textbf{BRKGA}    & $277.9$    & $197.2$   & $168.9$    & $335.3$    & $232.3$  &  $201.8$              \\
\textbf{BRKGA+}   & $271.1$    &  $186.4$  & $150.8$    & $308.1$    & $225.6$  &  $177.8$               \\
\bottomrule
\end{tabular}
\vspace{-4mm}
\end{table}

The proposed BRKGA method is suitable to solving both single-UAV and multi-UAV CPP problems. We first evaluated the proposed method with inspection tasks for two small-scale structures. A heuristic greedy neighborhood graph search (labeled as GNS\cite{jing2019coverage}), and a set covering based view planning with TSP path planning (labeled as VPP-TSP \cite{jing2016sampling, chen2004automatic}) were used as baselines for benchmarking. The reported BRKGA and BRKGA with local improvement heuristics (labeled as BRKGA+) results were the mean value based on 10 runs. Note that VPP-TSP is only applicable to the single-agent \gls{cpp} problem.

The parameters for BRKGA and BRKGA+ were as follows: the population size was 1000; the number of generation was 100; the mutation rate was 0.2, the crossover rate was 0.5; the elite selection rate was 0.1. For BRKGA+, the rate of performing local improvement heuristics was set as 0.2.

\begin{figure}[!ht]
    \centering
    \begin{subfigure}[b]{0.48\linewidth}
    	\includegraphics[width=0.99\linewidth]{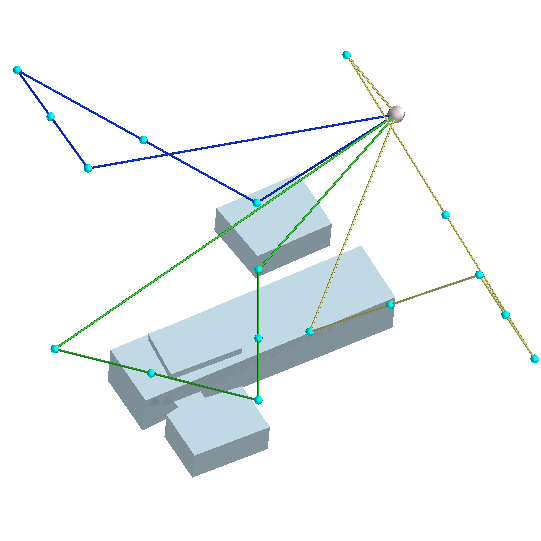}
    	\caption{T1}\label{fig:t1_traj}
    \end{subfigure}
        \begin{subfigure}[b]{0.48\linewidth}
    	\includegraphics[width=0.99\linewidth]{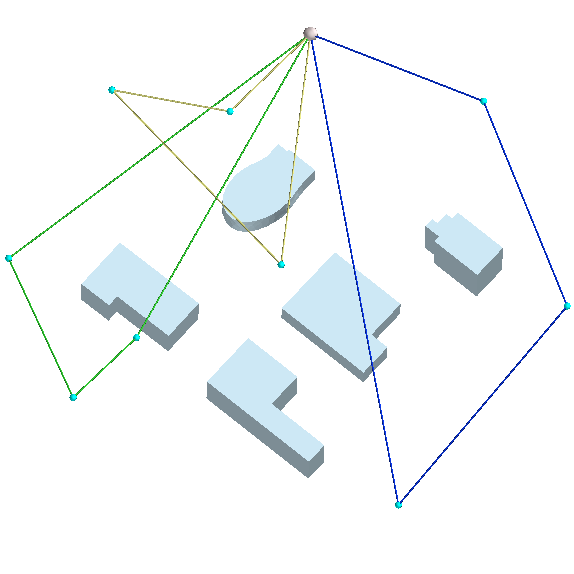}
    	\caption{T2}\label{fig:t2_traj}
    \end{subfigure}
    \caption{Example visualizations of planned paths with 3 UAVs, each color represents the path of an individual UAV }\label{fig:traj_t12}
    \vspace{-4mm}
\end{figure}

The results in Table \ref{table::res_1} show that the proposed \gls{brkga}-based method performs significantly better than previous methods. For the single UAV case, BRKGA reduces the path length by $44.1\% $ and  $31.4\% $ compared with the VPP-TSP and GNS methods, while BRKGA with local improvement heuristics (BRKGA+) further reduces it by $5.3\% $ over BRKGA and achieves an overall reduction of $47.1\% $ and  $35.1\%$. For the multi-UAV inspection path planning results, BRKGA and BRKGA+ reduce the path length, respectively, by $43.6\% $ and $48.0\% $ compared with GNS. The planned inspection paths are visualized in Fig. \ref{fig:traj_t12}, where each color represents the path of an individual UAV.

\begin{table*}[h]
\centering
\caption{Results of planned path for Multi-agent Inspection (measured in meters, average of 10 runs)} \label{table::res_2}
\begin{tabular}{ccccccccccccc}
\toprule
                & \multicolumn{3}{c}{\textbf{T3}} & \multicolumn{3}{c}{\textbf{T4}} & \multicolumn{3}{c}{\textbf{T5}} & \multicolumn{3}{c}{\textbf{T6}} \\
No. of UAVs     & 3         & 4        & 5        & 3         & 4        & 5        & 3         & 4        & 5        & 3         & 4        & 5        \\\midrule
\textbf{GNS\cite{jing2019coverage}}    &  $527.6$  &  $447.3$ & $430.5$  & $804.5$  & $542.5$  & $498.3$  & $1646.5$ & $1373.4$  & $1312.4$  & $1890.2$   & $1811.9$  & $1736.3$  \\
\textbf{BRKGA}  &  $396.1$  &  $351.3$ & $279.5$  & $697.2$  & $500.3$  & $449.8$  & $1417.7$ & $1167.8$  & $1017.8$  & $1821.4$   & $1529.9$  & $1490.4$   \\
\textbf{BRKGA+} &  $357.0$  &  $317.4$ & $270.6$  & $567.7$  & $430.9$  & $390.6$  & $1351.7$ & $1062.3$  & $926.7$   & $1783.5$   & $1349.2$  & $1321.8$  \\
\bottomrule
\end{tabular}
\vspace{-4mm}
\end{table*}

\subsection{Additional Results on Large-Scale, Complex Structures}

\begin{figure}[!ht]
    \centering
    \begin{subfigure}[b]{0.48\linewidth}
    	\includegraphics[width=0.99\linewidth]{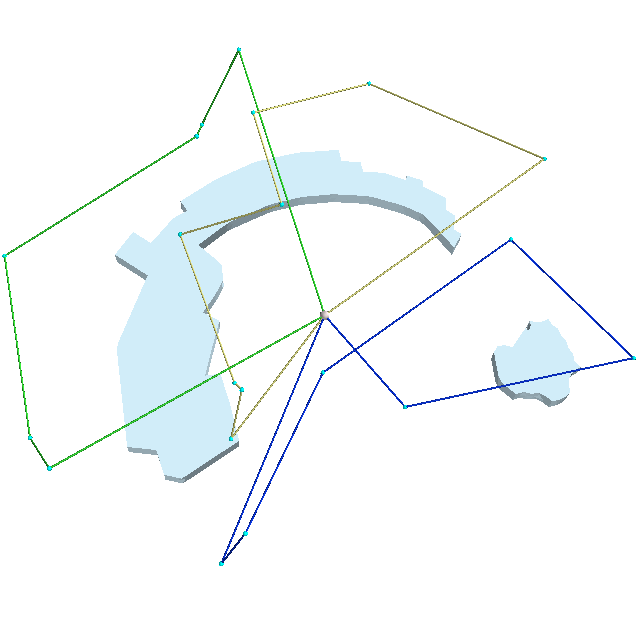}
    	\caption{T3}\label{fig:t3_traj}
    \end{subfigure}
        \begin{subfigure}[b]{0.48\linewidth}
    	\includegraphics[width=0.99\linewidth]{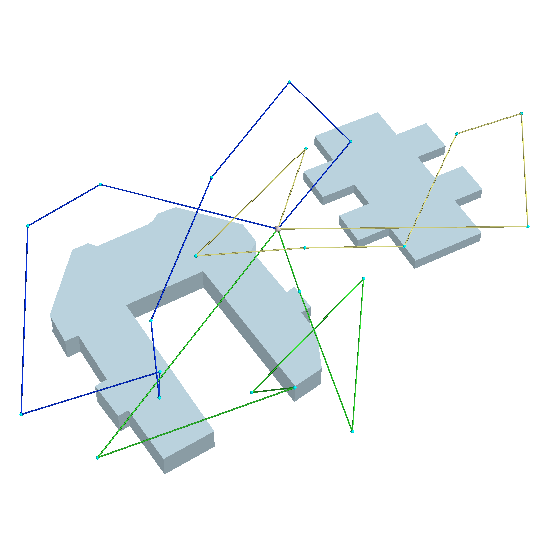}
    	\caption{T4}\label{fig:t4_traj}
    \end{subfigure}
    
    \begin{subfigure}[b]{0.48\linewidth}
    	\includegraphics[width=0.99\linewidth]{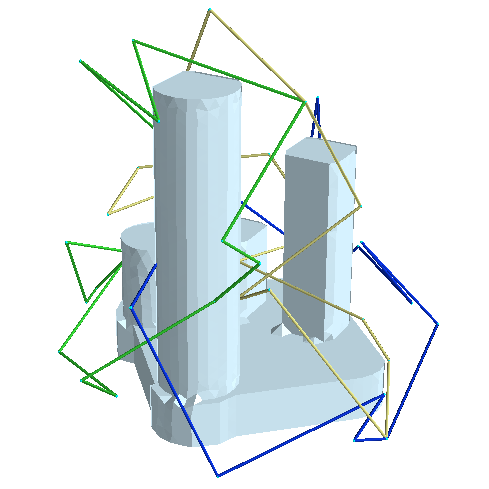}
    	\caption{T5}\label{fig:t5_traj}
    \end{subfigure}
        \begin{subfigure}[b]{0.48\linewidth}
    	\includegraphics[width=0.99\linewidth]{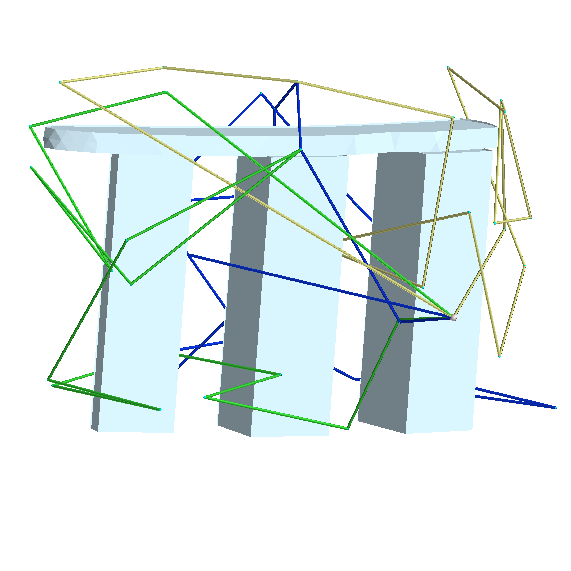}
    	\caption{T6}\label{fig:t6_traj}
    \end{subfigure}
    \caption{Example visualizations of planned paths with 3 UAVs, each color represents the path of an individual UAV.}\label{fig:traj_t3456}
    \vspace{-4mm}
\end{figure}

\noindent We also evaluated the performance of the proposed method with multiple UAVs for four additional large-scale, complex target structures. The population size was set to 2500 for BRKGA and BRKGA+, in order to deal with the larger scale problems and more UAVs. All the other parameters were the same as in Section \ref{sec:res_b}. In the planning results shown in Table \ref{table::res_2}, BRKGA and BRKGA+ reduced path lengths by $16.4\% $ and $24.6\% $ on average compared with the previous method. The planned inspection paths are visualized in Fig. \ref{fig:traj_t3456}.

\begin{figure}[!ht]
    \centering
    \begin{subfigure}[b]{0.49\linewidth}
    	\includegraphics[width=0.99\linewidth]{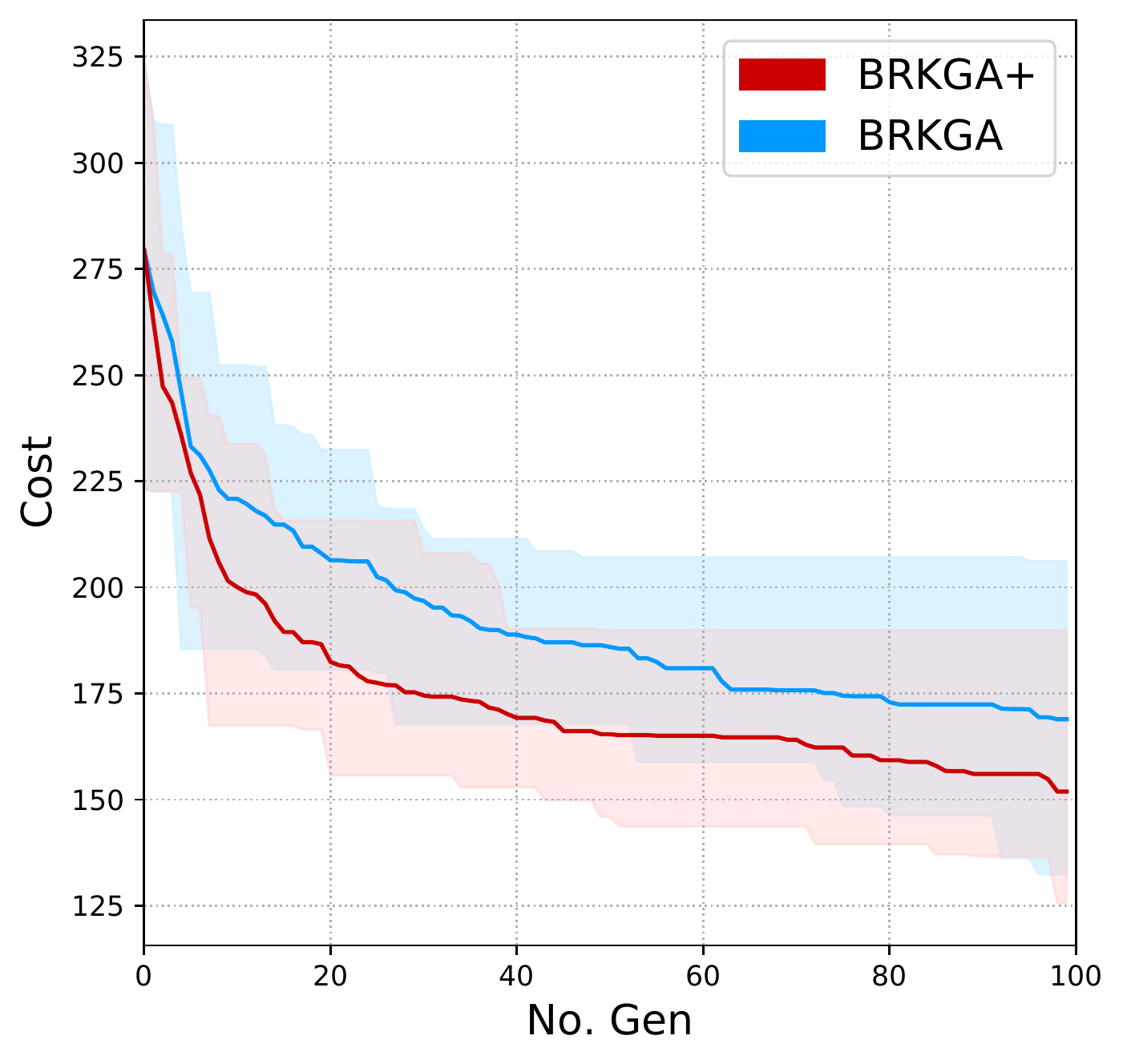}
    	\caption{Convergence on T1}\label{fig:rkga1}
    \end{subfigure}
    \begin{subfigure}[b]{0.49\linewidth}
    	\includegraphics[width=0.99\linewidth]{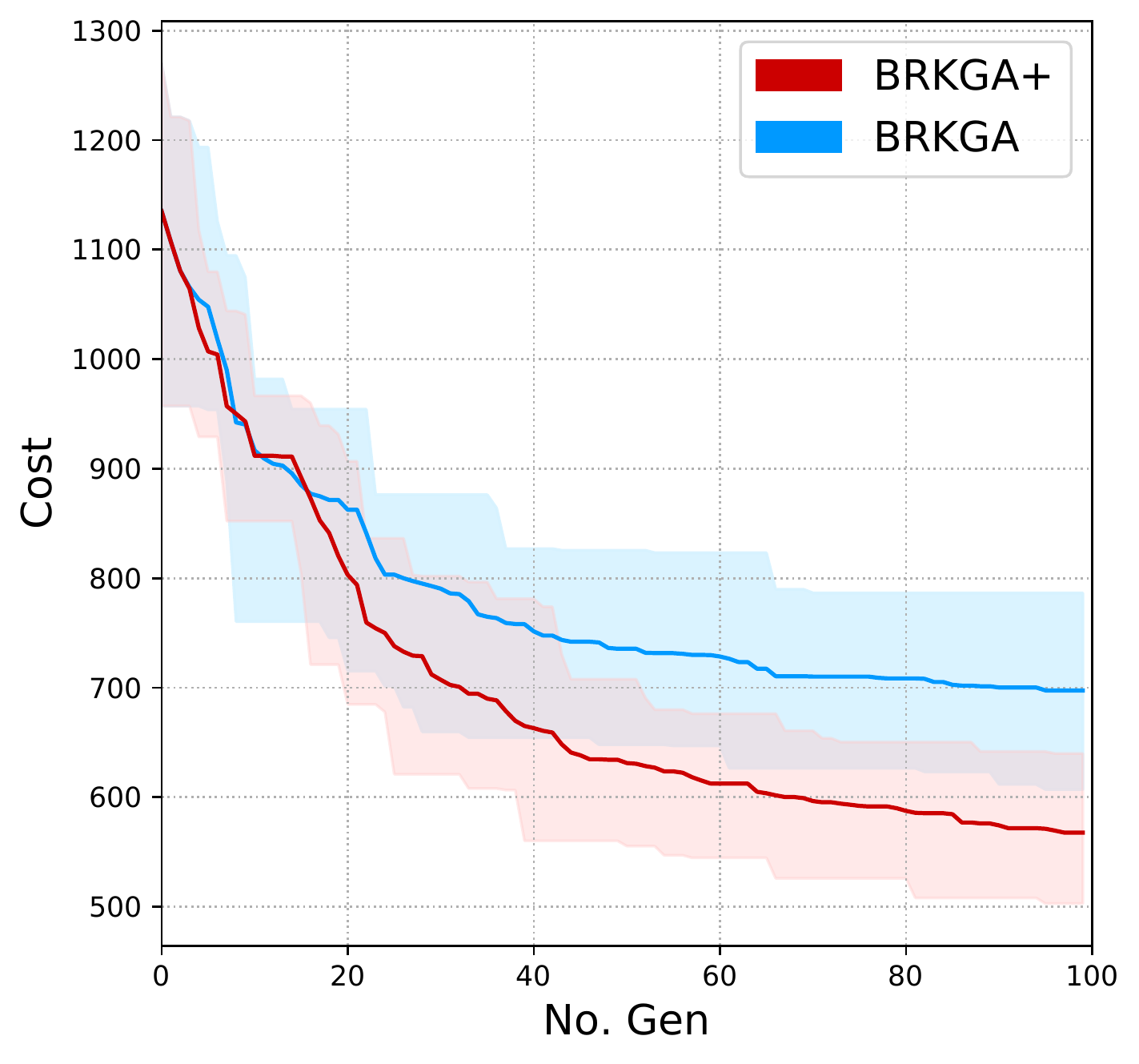}
    	\caption{Convergence on T4}\label{fig:rkga2}
    \end{subfigure}
    \caption{BRKGA convergence rates for three-UAV cases (based on 10-run)}\label{fig:rkga_conv}
    \vspace{-4mm}
\end{figure}

\begin{figure*}[h]
\centering
\begin{subfigure}[b]{0.28\linewidth}
	\includegraphics[width=0.99\linewidth]{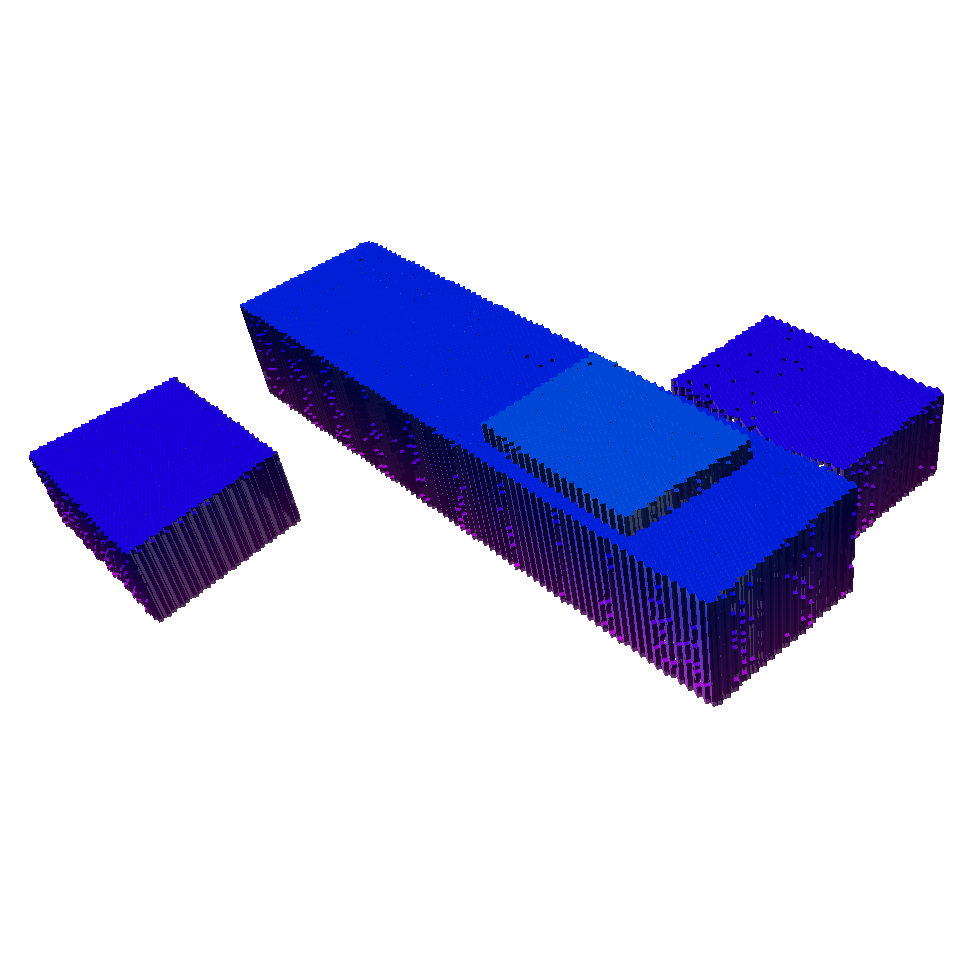}
	\caption{}\label{fig:vox_ny7}
\end{subfigure}    
\begin{subfigure}[b]{0.28\linewidth}
	\includegraphics[width=0.99\linewidth]{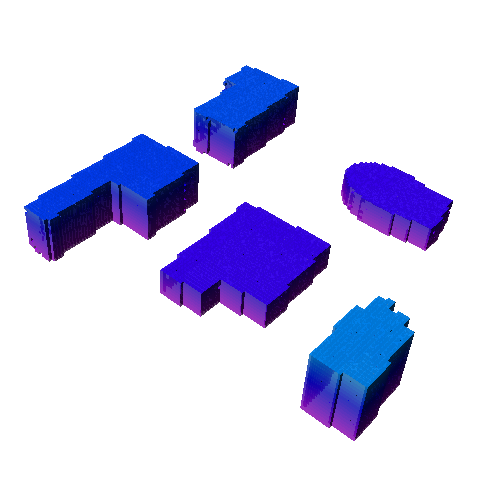}
	\caption{}\label{fig:vox_b42}
\end{subfigure}
\begin{subfigure}[b]{0.28\linewidth}
	\includegraphics[width=0.99\linewidth]{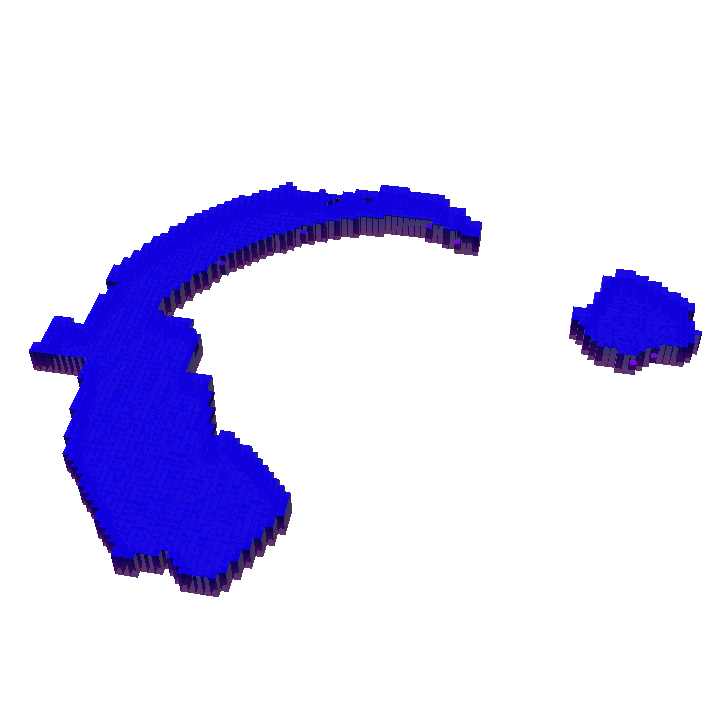}
	\caption{}\label{fig:vox_sentosa}
\end{subfigure} 

\begin{subfigure}[b]{0.28\linewidth}
	\includegraphics[width=0.99\linewidth]{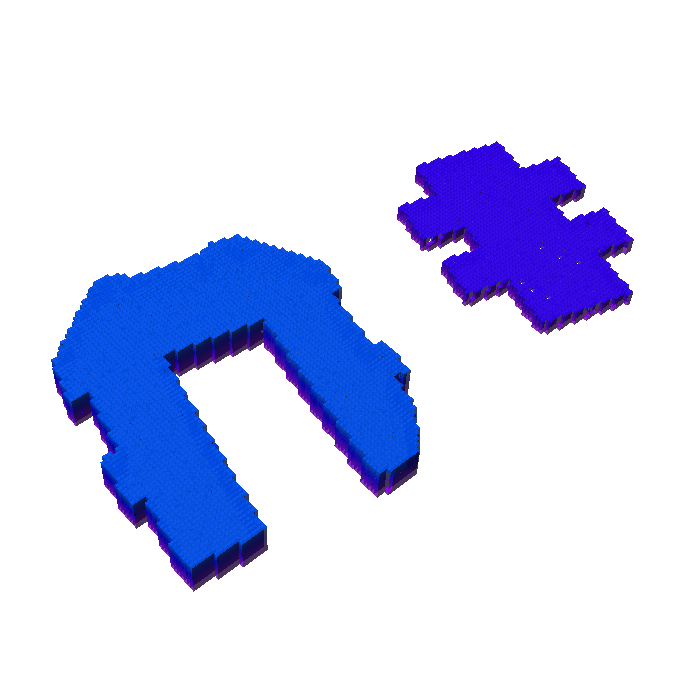}
	\caption{}\label{fig:vox_corp}
\end{subfigure}
\begin{subfigure}[b]{0.28\linewidth}
	\includegraphics[width=0.99\linewidth]{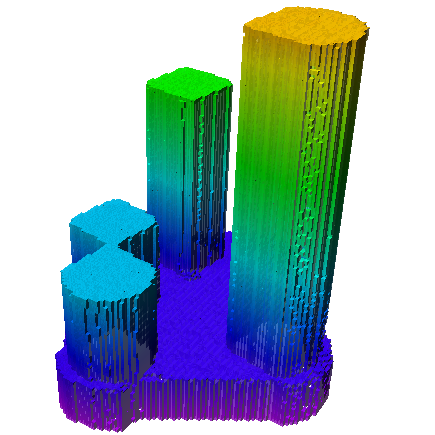}
	\caption{}\label{fig:vox_raffles}
\end{subfigure} 
\begin{subfigure}[b]{0.28\linewidth}
	\includegraphics[width=0.99\linewidth]{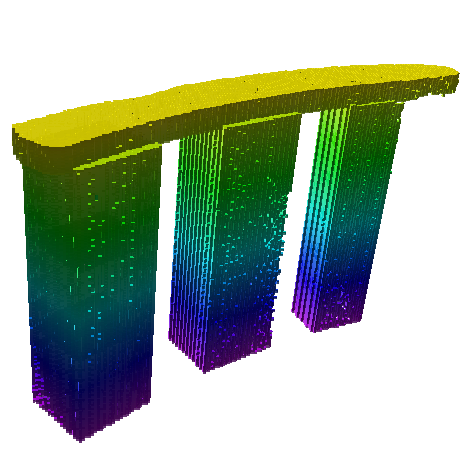}
	\caption{}\label{fig:vox_mbs}
\end{subfigure}
	\caption{Visualization of coverage target surfaces: the voxel model is obtained by simulating the planned path for three UAVs using the proposed method in Drake simulator}\label{fig:voxels}
	\vspace{-4mm}
\end{figure*}

The planned paths of multiple UAVs in different environments were simulated with Drake simulator \cite{drake}. Based on the simulated sensor measurements along the paths, Octomap \cite{hornung13auro} was used to incrementally construct the map and evaluate the quality of coverage. The example convergence plots of the BRKGA and BRKGA+ are shown in Fig. \ref{fig:rkga_conv}, which show both methods converge smoothly, but that BRKGA+ converges faster than BRKGA. The reconstructed octomaps are shown in Fig. \ref{fig:voxels}. 

\section{Discussion}

\noindent The proposed planning method automatically generates paths for multiple UAVs to inspect large-scale 3D structures with complex geometries. It also features a continuous 3D swapping coverage with the path-primitives, instead of discrete coverage with individual viewpoints. Our research demonstrated that there are several advantages of continuous 3D swapping coverage. First, the planned inspection paths are shorter than the discrete viewpoints planning method (VPP-TSP). Second, it provides more reliable image registration compared to viewpoint planning. 

Since the BRKGA method optimizes the inspection paths for multiple agents globally, it significantly outperformed the previous method. Additionally, increasing the number of UAVs reduces the cost (maximum path of the UAVs) of the inspection. Note, however, that the path length reduction is not proportional to the number of the UAVs, as also observed in other multi-agent coverage planning in 2D scenarios \cite{kim2020voronoi}.

The proposed method for finding efficient inspection paths is widely applicable to inspection applications for any geometries of complex 3D structure. Each individual UAV covers a separate area by following a unique path. Though not all planned paths are very smooth, especially for large-scale complex structures (e.g. T5, T6), the path smoothness issue usually happens on a long path-primitive, which is long enough for the UAV to follow. The path smooth issue could be addressed in future research by adding objectives/constraints in the problem formulation, since only the longest route inthe SC-VRP
determines the cost of the optimal solution, all other shorter routes are not optimized. In addition, the smoothness of the path could also be addressed by post path smoothing and optimization, or at lower control level through trajectory optimization while flying the UAVs along the planned path.

\section{Conclusion} \label{sec:conclusion}

\noindent The proposed multi-UAV path planning framework for inspecting large-scale, complex 3D structures consists of a coverage sampling phase for path-primitive dual sampling, and a path planning and optimization phase using BRKGA. We demonstrated that the proposed method automatically generates efficient collision-free inspection paths for different large-scale and complex 3D structures with a required coverage ratio. The proposed method found highly efficient paths that are up to $48\%$ shorter than those created previous methods. Future research of multi-UAV inspection path planning includes post path smoothing, optimization, and uncertainty handling during the deployment of actual UAV flights.

\section*{Acknowledgement}

\noindent This research is supported by A*STAR, Singapore, under its AME Programmatic Funding Scheme (Project \#A18A2b0046), and A*STAR Career Development Awards (CDA) (Project \#202D800028).

\bibliography{macpp2020}

\begin{thebibliography}{10}
\providecommand{\url}[1]{#1}
\csname url@rmstyle\endcsname
\providecommand{\newblock}{\relax}
\providecommand{\bibinfo}[2]{#2}
\providecommand\BIBentrySTDinterwordspacing{\spaceskip=0pt\relax}
\providecommand\BIBentryALTinterwordstretchfactor{4}
\providecommand\BIBentryALTinterwordspacing{\spaceskip=\fontdimen2\font plus
\BIBentryALTinterwordstretchfactor\fontdimen3\font minus
  \fontdimen4\font\relax}
\providecommand\BIBforeignlanguage[2]{{%
\expandafter\ifx\csname l@#1\endcsname\relax
\typeout{** WARNING: IEEEtran.bst: No hyphenation pattern has been}%
\typeout{** loaded for the language `#1'. Using the pattern for}%
\typeout{** the default language instead.}%
\else
\language=\csname l@#1\endcsname
\fi
#2}}

\bibitem{bircher2015structural}
A.~Bircher, K.~Alexis, M.~Burri, P.~Oettershagen, S.~Omari, T.~Mantel, and
  R.~Siegwart, ``Structural inspection path planning via iterative viewpoint
  resampling with application to aerial robotics,'' in \emph{IEEE International
  Conference on Robotics and Automation}.\hskip 1em plus 0.5em minus
  0.4em\relax IEEE, 2015, pp. 6423--6430.

\bibitem{jing2016sampling}
W.~Jing, J.~Polden, W.~Lin, and K.~Shimada, ``Sampling-based view planning for
  3d visual coverage task with unmanned aerial vehicle,'' in \emph{IEEE/RSJ
  International Conference on Intelligent Robots and Systems}.\hskip 1em plus
  0.5em minus 0.4em\relax IEEE, 2016, pp. 1808--1815.

\bibitem{roberts2017submodular}
M.~Roberts, S.~Shah, D.~Dey, A.~Truong, S.~N. Sinha, A.~Kapoor, P.~Hanrahan,
  and N.~Joshi, ``Submodular trajectory optimization for aerial 3d scanning.''
  in \emph{International Conference on Computer Vision (ICCV)}, 2017, pp.
  5334--5343.

\bibitem{almadhoun2019survey}
R.~Almadhoun, T.~Taha, L.~Seneviratne, and Y.~Zweiri, ``A survey on multi-robot
  coverage path planning for model reconstruction and mapping,'' \emph{SN
  Applied Sciences}, vol.~1, no.~8, p. 847, 2019.

\bibitem{OpenStreetMap}
{OpenStreetMap contributors}, ``{Planet dump retrieved from
  https://planet.osm.org },'' \url{ https://www.openstreetmap.org }, 2017.

\bibitem{almadhoun2016survey}
R.~Almadhoun, T.~Taha, L.~Seneviratne, J.~Dias, and G.~Cai, ``A survey on
  inspecting structures using robotic systems,'' \emph{International Journal of
  Advanced Robotic Systems}, vol.~13, no.~6, p. 1729881416663664, 2016.

\bibitem{scott2003view}
W.~Scott, G.~Roth, and J.-F. Rivest, ``View planning for automated 3d object
  reconstruction inspection,'' \emph{ACM Computing Surveys}, vol.~35, no.~1,
  2003.

\bibitem{hornung13auro}
\BIBentryALTinterwordspacing
A.~Hornung, K.~M. Wurm, M.~Bennewitz, C.~Stachniss, and W.~Burgard,
  ``{OctoMap}: An efficient probabilistic {3D} mapping framework based on
  octrees,'' \emph{Autonomous Robots}, 2013, software available at
  \url{http://octomap.github.com}. [Online]. Available:
  \url{http://octomap.github.com}
\BIBentrySTDinterwordspacing

\bibitem{song2017online}
S.~Song and S.~Jo, ``Online inspection path planning for autonomous 3d modeling
  using a micro-aerial vehicle,'' in \emph{2017 IEEE International Conference
  on Robotics and Automation (ICRA)}.\hskip 1em plus 0.5em minus 0.4em\relax
  IEEE, 2017, pp. 6217--6224.

\bibitem{jing2019coverage}
W.~Jing, D.~Deng, Z.~Xiao, Y.~Liu, and K.~Shimada, ``Coverage path planning
  using path primitive sampling and primitive coverage graph for visual
  inspection,'' in \emph{IEEE/RSJ International Conference on Intelligent
  Robots and Systems}.\hskip 1em plus 0.5em minus 0.4em\relax IEEE, 2019, pp.
  1472--1479.

\bibitem{cabreira2019survey}
T.~M. Cabreira, L.~B. Brisolara, and P.~R. Ferreira~Jr, ``Survey on coverage
  path planning with unmanned aerial vehicles,'' \emph{Drones}, vol.~3, no.~1,
  p.~4, 2019.

\bibitem{deng2019constrained}
D.~Deng, W.~Jing, Y.~Fu, Z.~Huang, J.~Liu, and K.~Shimada, ``Constrained
  heterogeneous vehicle path planning for large-area coverage,'' in
  \emph{IEEE/RSJ International Conference on Intelligent Robots and
  Systems}.\hskip 1em plus 0.5em minus 0.4em\relax IEEE, 2019, pp. 4113--4120.

\bibitem{lavalle2006planning}
S.~M. LaValle, \emph{Planning algorithms}.\hskip 1em plus 0.5em minus
  0.4em\relax Cambridge university press, 2006.

\bibitem{choset2001coverage}
H.~Choset, ``Coverage for robotics--a survey of recent results,'' \emph{Annals
  of mathematics and artificial intelligence}, vol.~31, no. 1-4, pp. 113--126,
  2001.

\bibitem{fu2019toward}
M.~Fu, A.~Kuntz, O.~Salzman, and R.~Alterovitz, ``Toward asymptotically-optimal
  inspection planning via efficient near-optimal graph search,'' \emph{arXiv
  preprint arXiv:1907.00506}, 2019.

\bibitem{vasquez2019coverage}
J.~I. Vasquez-Gomez, M.~Marciano-Melchor, L.~Valentin, and J.~C.
  Herrera-Lozada, ``Coverage path planning for 2d convex regions,''
  \emph{Journal of Intelligent \& Robotic Systems}, pp. 1--14, 2019.

\bibitem{hepp2018plan3d}
B.~Hepp, M.~Nie{\ss}ner, and O.~Hilliges, ``Plan3d: Viewpoint and trajectory
  optimization for aerial multi-view stereo reconstruction,'' \emph{ACM
  Transactions on Graphics (TOG)}, vol.~38, no.~1, pp. 1--17, 2018.

\bibitem{englot2013three}
B.~Englot and F.~S. Hover, ``Three-dimensional coverage planning for an
  underwater inspection robot,'' \emph{The International Journal of Robotics
  Research}, vol.~32, no. 9-10, pp. 1048--1073, 2013.

\bibitem{scott2009model}
W.~R. Scott, ``Model-based view planning,'' \emph{Machine Vision and
  Applications}, vol.~20, no.~1, pp. 47--69, 2009.

\bibitem{jing2016view}
W.~Jing, J.~Polden, P.~Y. Tao, W.~Lin, and K.~Shimada, ``View planning for 3d
  shape reconstruction of buildings with unmanned aerial vehicles,'' in
  \emph{International Conference on Control, Automation, Robotics and
  Vision}.\hskip 1em plus 0.5em minus 0.4em\relax IEEE, 2016, pp. 1--6.

\bibitem{kaba2017reinforcement}
M.~D. Kaba, M.~G. Uzunbas, and S.-N. Lim, ``A reinforcement learning approach
  to the view planning problem.'' in \emph{Conference on Computer Vision and
  Pattern Recognition (CVPR)}, 2017, pp. 5094--5102.

\bibitem{galceran2013survey}
E.~Galceran and M.~Carreras, ``A survey on coverage path planning for
  robotics,'' \emph{Robotics and Autonomous Systems}, vol.~61, no.~12, pp.
  1258--1276, 2013.

\bibitem{nedjati2016complete}
A.~Nedjati, G.~Izbirak, B.~Vizvari, and J.~Arkat, ``Complete coverage path
  planning for a multi-uav response system in post-earthquake assessment,''
  \emph{Robotics}, vol.~5, no.~4, p.~26, 2016.

\bibitem{kim2020voronoi}
J.~Kim and H.~I. Son, ``A voronoi diagram-based workspace partition for weak
  cooperation of multi-robot system in orchard,'' \emph{IEEE Access}, 2020.

\bibitem{choi2019energy}
Y.~Choi, Y.~Choi, S.~Briceno, and D.~N. Mavris, ``Energy-constrained multi-uav
  coverage path planning for an aerial imagery mission using column
  generation,'' \emph{Journal of Intelligent \& Robotic Systems}, pp. 1--15,
  2019.

\bibitem{mansouri2018cooperative}
S.~S. Mansouri, C.~Kanellakis, E.~Fresk, D.~Kominiak, and G.~Nikolakopoulos,
  ``Cooperative coverage path planning for visual inspection,'' \emph{Control
  Engineering Practice}, vol.~74, pp. 118--131, 2018.

\bibitem{renzaglia2019multi}
A.~Renzaglia, J.~Dibangoye, V.~L. Doze, and O.~Simon, ``Multi-uav visual
  coverage of partially known 3d surfaces: Voronoi-based initialization to
  improve local optimizers,'' \emph{arXiv preprint arXiv:1901.10272}, 2019.

\bibitem{chen2011active}
S.~Chen, Y.~Li, and N.~M. Kwok, ``Active vision in robotic systems: A survey of
  recent developments,'' \emph{The International Journal of Robotics Research},
  vol.~30, no.~11, pp. 1343--1377, 2011.

\bibitem{tarbox1995planning}
G.~H. Tarbox and S.~N. Gottschlich, ``Planning for complete sensor coverage in
  inspection,'' \emph{Computer Vision and Image Understanding}, vol.~61, no.~1,
  pp. 84--111, 1995.

\bibitem{kavraki1996probabilistic}
L.~E. Kavraki, P.~{\v{S}}vestka, J.-C. Latombe, and M.~H. Overmars,
  ``Probabilistic roadmaps for path planning in high-dimensional configuration
  spaces,'' \emph{IEEE TRANSACTIONS ON ROBOTICS AND AUTOMATION}, vol.~12,
  no.~4, 1996.

\bibitem{dantzig1954solution}
G.~Dantzig, R.~Fulkerson, and S.~Johnson, ``Solution of a large-scale
  traveling-salesman problem,'' \emph{Journal of the operations research
  society of America}, vol.~2, no.~4, pp. 393--410, 1954.

\bibitem{laporte1992traveling}
G.~Laporte, ``The traveling salesman problem: An overview of exact and
  approximate algorithms,'' \emph{European Journal of Operational Research},
  vol.~59, no.~2, pp. 231--247, 1992.

\bibitem{bean1994genetic}
J.~C. Bean, ``Genetic algorithms and random keys for sequencing and
  optimization,'' \emph{ORSA journal on computing}, vol.~6, no.~2, pp.
  154--160, 1994.

\bibitem{snyder2006random}
L.~V. Snyder and M.~S. Daskin, ``A random-key genetic algorithm for the
  generalized traveling salesman problem,'' \emph{European journal of
  operational research}, vol. 174, no.~1, pp. 38--53, 2006.

\bibitem{gonccalves2011biased}
J.~F. Gon{\c{c}}alves and M.~G. Resende, ``Biased random-key genetic algorithms
  for combinatorial optimization,'' \emph{Journal of Heuristics}, vol.~17,
  no.~5, pp. 487--525, 2011.

\bibitem{jing2017redundant}
W.~Jing, J.~Polden, C.~F. Goh, M.~Rajaraman, W.~Lin, and K.~Shimada,
  ``Sampling-based coverage motion planning for industrial inspection
  application with redundant robotic system,'' in \emph{IEEE/RSJ International
  Conference on Intelligent Robots and Systems}.\hskip 1em plus 0.5em minus
  0.4em\relax IEEE, 2017, pp. 5211--5218.

\bibitem{croes1958method}
G.~A. Croes, ``A method for solving traveling-salesman problems,''
  \emph{Operations research}, vol.~6, no.~6, pp. 791--812, 1958.

\bibitem{DEAP_JMLR2012}
F.-A. Fortin, F.-M. {De Rainville}, M.-A. Gardner, M.~Parizeau, and C.~Gagn\'e,
  ``{DEAP}: Evolutionary algorithms made easy,'' \emph{Journal of Machine
  Learning Research}, vol.~13, pp. 2171--2175, jul 2012.

\bibitem{hagberg2008exploring}
A.~Hagberg, P.~Swart, and D.~S~Chult, ``Exploring network structure, dynamics,
  and function using networkx,'' Los Alamos National Lab.(LANL), Los Alamos, NM
  (United States), Tech. Rep., 2008.

\bibitem{chen2004automatic}
S.~Chen and Y.~Li, ``Automatic sensor placement for model-based robot vision,''
  \emph{IEEE Transactions on Systems, Man, and Cybernetics, Part B:
  Cybernetics}, vol.~34, no.~1, pp. 393--408, 2004.

\bibitem{drake}
\BIBentryALTinterwordspacing
R.~Tedrake and the Drake Development~Team, ``Drake: Model-based design and
  verification for robotics,'' 2019. [Online]. Available:
  \url{https://drake.mit.edu}
\BIBentrySTDinterwordspacing

\end{thebibliography}

\end{document}